\documentclass[10pt,twocolumn,letterpaper]{article}

\usepackage{cvpr}              

\usepackage{graphicx}
\usepackage{amsmath}
\usepackage{amssymb}
\usepackage{booktabs}
\usepackage{verbatim}
\usepackage{subcaption}
\usepackage{float}
\usepackage{amsfonts}
\usepackage{ulem}
\newtheorem{theorem}{Theorem}
\newtheorem{lemma}[theorem]{Lemma}
\newtheorem{hyp}{Hypothesis}
\usepackage{ulem}

\usepackage[pagebackref,breaklinks,colorlinks]{hyperref}

\usepackage[capitalize]{cleveref}
\crefname{section}{Sec.}{Secs.}
\Crefname{section}{Section}{Sections}
\Crefname{table}{Table}{Tables}
\crefname{table}{Tab.}{Tabs.}

\def\cvprPaperID{8392} 
\def\confName{CVPR}
\def\confYear{2022}

\begin{document}

\title{Couplformer: Rethinking Vision Transformer with Coupling Attention Map}

\author{Hai Lan\\
Fujian Institute of Research on the Structure of Matter\\
Chinese Academy of Sciences\\
{\tt\small lanhai09@fjirsm.ac.cn}
\and
Xihao Wang$^*$\\
Technical University of Munich\\
{\tt\small xihaowang2016@gmail.com}
\thanks{Equal technical contribution}
\and
Xian Wei$^\dag $ \\
East China Normal University\\
{\tt\small xian.wei@tum.de} 
\thanks{Corresponding Author}}

\maketitle
\begin{abstract}
With the development of the self-attention mechanism, the Transformer model has demonstrated its outstanding performance in the computer vision domain. 
However, the massive computation brought from the full attention mechanism became a heavy burden for memory consumption. Sequentially, the limitation of memory reduces the possibility of improving the Transformer model. 
To remedy this problem, we propose a novel memory economy attention mechanism named Couplformer, which decouples the attention map into two sub-matrices and generates the alignment scores from spatial information. A series of different scale image classification tasks are applied to evaluate the effectiveness of our model. The result of experiments shows that on the ImageNet-1k classification task, the Couplformer can significantly decrease $28\%$ memory consumption compared with regular Transformer while accessing sufficient accuracy requirements and outperforming $0.92\%$ on Top-1 accuracy while occupying the same memory footprint. 
As a result, the Couplformer can serve as an efficient backbone in visual tasks, and provide a novel perspective on the attention mechanism for researchers.
\end{abstract}

\section{Introduction}
\label{seq:01}

In recent years, attention mechanism is generating considerable interest in the domain of deep learning.
The main breakthroughs of attention modules were first appeared in Natural Language Processing (NLP) \cite{attentionisallyouneed,bigbird,reformer,memformer,CompreTrans}, and then also in computer vision domain. \cite{ViT16times16,detectdeformable,huang2019attention_AOA_iccv,FAHIM2020106437,swintrans}. These achievements have demonstrated that the attention module provided a different approach than CNN models to handle the task and demonstrated a promising performance.

\begin{figure}[t]
	\centering
    \includegraphics[width=0.48\textwidth]{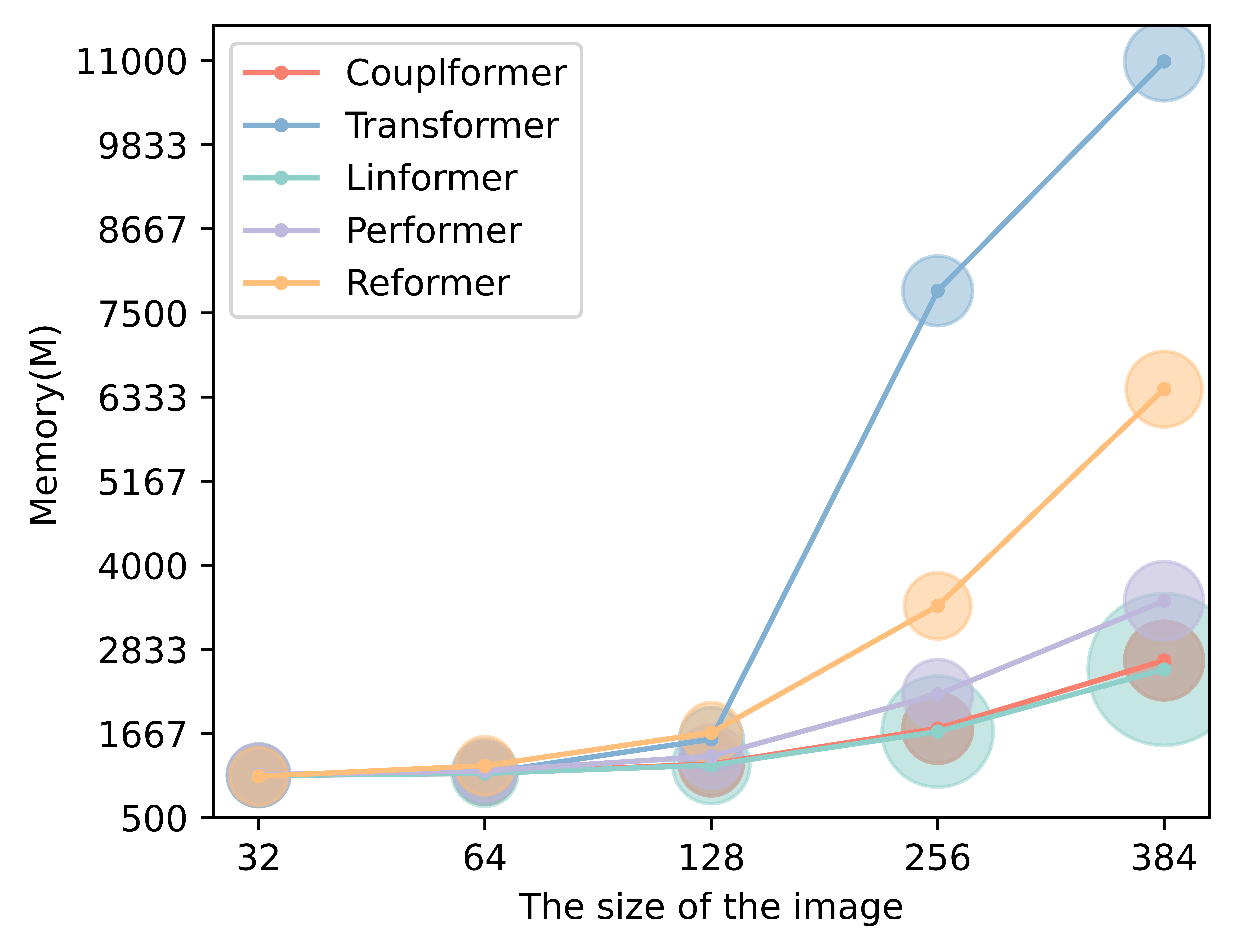}
	\caption{The proposed Couplformer achieves memory economy in the visual task. When the patch size is unvaried, the memory consumption increases with the growth of the image size. Compared with the original Transformer, efficient Transformer models present a lower memory consumption to a relative degree. Our Couplformer keeps the most economy memory increment with the minimum training parameters. (the scatter area means the number of training parameters).}
	\label{fig:memory}
\end{figure}


Generally speaking, the attention module tends to exert a learnable weight on features to distinguish importance from various perspectives. As a particular form of attention, self-attention is applied as the core mechanism of the neural network, named Transformer. Initially, the Transformer\cite{attentionisallyouneed} was first proposed to overcome the challenge of long-range dependency in large machine translation tasks. Following the success in the NLP domain, the Vision Transformer\cite{ViT16times16} also demonstrates its powerful ability in handling the large-scale vision dataset. The essential advantage of self-attention is that it focuses on the relative importance of own content rather than weighting across multiple contents as a general attention mechanism. Nowadays, Transformer and attention-based models have become ubiquitous in the modern machine learning area.

Unfortunately, there are a few drawbacks of the Transformer model.
Coming with outstanding performance, it also exposed the severe demand for computation and memory. Because the full self-attention is the quadratic dependency on the sequence length, the time and space complexity extend considerably\cite{sparformer,longformer,reformer,bigbird,linformer}. Several challenges result from this problem. First of all, compared with the ConvNets, the larger memory consumption of Transformer restricts the processing of large-scale datasets. It hinders the Transformer from performing its unique transferability from large-scale datasets to smaller-scale datasets. Secondly, regarding the Vision Transformer, it requires leveraging the patch embeddings to encode the small region of the image into a single input token\cite{ViT16times16}. Due to the limitation of memory consumption, the selection of patch size tends to be larger in order to reduce the length of the token sequence. However, patch representation is a critical component to the performance of the Transformer\cite{mlpmixer}. Moreover, with the growth of the image size, the problem of memory limitation will be much more conspicuous. In a word, memory consumption inhibits the Transformer model scalability in quite a lot of settings.   

In this paper, we introduce a novel method, Couplformer, which involves the coupling attention mechanism to alleviate the memory limitation in Vision Transformers. From the \cref{fig:memory}, Couplformer merely requires 31.6\% memory consumption of original Vision Transformer(ViT). Although other efficient Transformers\cite{linformer,sparformer,longformer,reformer} employs partial attention scores to replace the full attention matrix to break the memory limitation, the expression ability of the model decreases due to the low-rank bottleneck\cite{lowrank}. But in our method, the coupling attention mechanism not only reduces the memory consumption but also keeps the same rank as the full-attention matrix. To evaluate our method, we elaborate a framework for image classification based on Compact Convolutional Transformer\cite{cct}. And the experiment results show Couplformer could be applied to different scales of datasets and obtain competitive performance by training from scratch. 

The main contributions of this paper include:
\begin{enumerate}
	\item  We introduced coupling attention mechanism which decoupled the self-attention map into two sub-matrices. And we design a novel way to construct the sub-matrix by calculating the alignment score between the vectors along the height and width axis rather than the channel axis.
	\item  We proposed our model, named Couplformer, which efficiently calculates the alignment scores from spatial information by leveraging the vectorization trick. The time and space complexity of our approach is reduced to $\mathcal{O}(kn)$. Especially, our Couplformer model conspicuous decreases the memory consumption and alleviates the crisis of limited computation resources. 
	\item To evaluate our model, we conduct different scales of image classification tasks using Couplformer without bells and whistles. According to the experiments, our model achieves competitive accuracy on different datasets with relatively low memory consumption.  
\end{enumerate}

\section{Related Work}

The self-attention attracts considerable interest due to its versatile applications. For CNN-based models, self-attention mechanism has been used in many modules,
such as extra re-weighting modules for channels \cite{gatherexcite,SEnet,residualatt}, jointly spatial and channel attention module \cite{chen2017sca_cvpr,cbam,dualatt},
or remold convolution operation with self-attention \cite{bello2019attention_augment_iccv,standalone,Cordonnier2020On_relation_iclr,zhao2020exploring_CVPR}.

Another line of works arrange self-attention to be a component in a pipeline for downstream tasks, 
such as augmentation of feature maps for classification \cite{bello2019attention_augment_iccv}, object detection \cite{detectdeformable,relationOD,endtoendOD}, and segmentation \cite{axialsegmentation}. 
However, it is difficult to directly apply the self-attention mechanism to pixel-wise data.
\cite{locality} restricted self-attention convolution within local neighborhoods for query pixels,
and \cite{axialsegmentation,axialtransformer} restricted computation along individual axes. 
The work \cite{igpt} reduced image resolution and color space before Transformer, and \cite{ViT16times16} directly applied Transformer on image sequence patches. 

\subsection{Vision Transformer}
\label{sec:0201}

Before \textit{Dosoviskiy et al}.\cite{ViT16times16} proposed the Vision Transformer, CNNs dominated the area of visual recognition. Depending on the weight sharing, scale separation, and shift equivariant, CNNs possess the powerful and efficient ability to extract the feature from the image\cite{cnndeeplearn}. Although the Transformer network didn't present the strong equivariance representation as CNNs, its unique structure endows its permutation equivariance to obtain inductive bias\cite{permuteequivaraint}.  

In detail, the standard Visual Transformer's structure includes the following parts: Token Embedding, Positional Embedding, Transformer Encoder, and Classification Head\cite{vitsurvey}. With the exploration of the Transformer network, various Transformer-based models were proposed to solve the vision tasks efficiently. \textit{Touvron et al}.\cite{DeiT} proposed the DeiT, which uses distillation learning to overcome the drawback of ViT that it could only present the outstanding performance in large-scale datasets. CPVT\cite{CPVT} model applied the different position embedding methods to improve the efficiency and flexibility of the ViT model. According to the CvT\cite{CVT} and CeiT\cite{CeiT} model, they try to hybridize the Transformer and CNN network to derive desirable properties from each one. \textit{Liu et al}.\cite{swintrans} proposed the Swin Transformer, which adopts a series of approaches in terms of visual tasks, such as patch partition, linear embedding, pyramid structure, and window-based MSA.   

\subsection{Efficient Transformers}
\label{sec:0202}

Depending on the self-attention mechanism, the Transformer model has already become prevalent in many fields. As described above, the standard self-attention operation relies on dot-production multiplication. Sequentially, this leads to the problem that self-attention is the quadratic time and memory complexity\cite{efficientsurvey}. Moreover, when the input sequence length grows, the poorly scales become the bottleneck in Transformer. The reason is that the dot product between the feature matrix $Q$ and matrix $K$ generates a massive matrix to present the token-token interaction. In such a situation, it is unavoidable to exist exhaustive and redundant computation in the standard self-attention operation.  

In order to address this problem, the researchers proposed several novel Transformer architectures to improve the original self-attention mechanism. Because all of them attempt to reduce the computational costs to let the model to perform efficiently, these models are named "efficient Transformers"\cite{efficientsurvey}. The complexity of efficient Transformers are listed in \cref{tab:complexity}. These approaches could be divided into three categories. 

The first solution is to sparsify the self-attention layers. According to the sparse attention, the attention map can only be computed by limited pairs in a particular pre-defined manner. For example, By limiting and fixing the field of view, \cite{blockwise} and \cite{locality} employed the fixed block local attention patterns to constrain the pair for the score. Similarly, \cite{sparformer} and \cite{longformer} leverage fixed strided attention patterns to achieve the cost reduction. \textit{Kitaev et al}.\cite{reformer} proposed a learnable approach by using hash-based similarity to replace the token-to-token interaction. Moreover, \textit{Zaheer et al}.\cite{bigbird} proposed their sparsity pattern, which is a universal approximation and Turing completeness. 

\begin{table}[H]\small
    \centering
    \begin{tabular}{c|c}
    \toprule
    Model & Complexity\\
    \midrule
    ViT\cite{ViT16times16} & $\mathcal{O}(n^{2})$ \\
	Reformer\cite{reformer} & $\mathcal{O}(nlog(n))$ \\
    Longformer\cite{longformer} & $\mathcal{O}(n(k+m))$ \\
	Linformer\cite{linformer} &  $\mathcal{O}(kn)$\\
	Performer\cite{performer} & $\mathcal{O}(kn)$\\
	Big Bird\cite{bigbird} & $\mathcal{O}(kn)$\\
	Synthesizer\cite{synthesizer} & $\mathcal{O}(n^{2})$\\
	Sparse Transformer\cite{sparformer} & $\mathcal{O}(n\sqrt{n})$ \\
	Transformer-XL\cite{transX}& $\mathcal{O}(n^{2})$\\
	Compression Transformer\cite{CompreTrans} & $\mathcal{O}(n^{2})$\\
	Ours & $\mathcal{O}(kn)$\\
	\bottomrule
    \end{tabular}
    \caption{Complexity of Efficient Transformer. n denotes the sequence length, k and m denote the arbitrary integer.}
    \label{tab:complexity}
\end{table}

Secondly, depending on the low rank prior, employing the kernel-based method to approximate the attention matrix could also reduce the complexity. In terms of approximation solution, \cite{linformer},\cite{performer} and \cite{linearTransformer} utilized the kernel method to avoid explicitly implementing the dot production. They attempted to find a relatively low-rank structure to reduce the memory and computational complexity. \textit{Tay et al.}\cite{synthesizer} proposed the Synthesizer Transformer, which leverages Multi-Layer Perception(MLP) structure to approximate the dot-product multiplication of self-attention mechanism. 

Lastly, there are some of the other efficient Transformer architectures different from the solutions mentioned above. Depending on the segment-based recurrence, \cite{transX} applies a hidden state to connect adjacent blocks with a recurrent mechanism\cite{efficientsurvey}. Different from the \cite{transX}, \cite{CompreTrans} utilizes a dual memory system to maintain a fine-grained memory of past segment activations. As one kind of efficient Transformer, our model could be classified into the approximation solution to reduce the computation complexity and memory consumption.  
\section{Couplformer}
\label{sec:3}

In this section, a brief review of self-attention mechanism is given firstly, then we describe the main idea of coupling attention and its efficient calculation. Finally, we introduce our elaboration of Couplformer model for image classification.

\subsection{Standard Attention Mechanism}
Let $L$ be the length of input tokens sequence, the 2D feature maps of image are converted to the 1D token sequences by raster-scanned method, hence $L = h*w $ while $h$ denotes the height and $w$ denotes the width of input 2D feature map. Three linear transformations are applied on the input tokens to generate the $Q,K,V \in \mathbb{R}^{L \times d}$ respectively. The output of self-attention module $O \in \mathbb{R}^{L \times d}$ can be calculated as below:
\begin{equation}
    AM = \frac{Q \cdot K^\top}{\sqrt{d}},  O = SM(AM) \cdot V  \label{eq:vanilla_attention}
\end{equation}
Here $SM(\cdot)$ denotes the softmax operation along the matrix's rows. $AM \in \mathbb{R}^{L \times L}$ denotes the attention matrix, which has $O(L^2)$ space complexity. Intuitively, the quadratic dependency of the attention matrix leads to high memory consumption and limits the application of the large feature maps in the CV scenario.

\subsection{Coupling Attention Map}

Inspired by the idea of applying decomposition or dimension reduction to attention matrix to reduce the complexity of attention mechanism \cite{DBLP:conf/aaai/XiongZCTFLS21,linformer,reformer}, We assume the attention map $AW \in \mathbb{R}^{hw \times hw}$ can be approximately decoupled into two sub-matrix $A \in \mathbb{R}^{h \times h}$ and $B \in \mathbb{R}^{w \times w}$.

\begin{hyp} 
\label{hyp:appro}
$$ AM \approx \widehat{AM} =  A \otimes B $$ 
\end{hyp}

Here $\otimes$ denotes the tensor product operator, which can be implemented by Kronecker product while $A$ and $B$ are two matrices, explicitly the Kronecker product of $\mathbf{A}$ and $\mathbf{B}$ is defined by:
\begin{equation}
	\label{eq:Kronecker}
	\textbf{A} \otimes \textbf{B}=
	\begin{bmatrix}
		a_{0,0}\textbf{B} & \cdots & a_{0,w-1}\textbf{B}\\
		\vdots & \ddots & \vdots\\
		a_{h-1,0}\textbf{B} & \cdots & a_{h-1,h-1}\textbf{B}\\
	\end{bmatrix}
\end{equation}

The lowercase $a_{*,*}$ denotes the element in the matrix, and the indexes of matrix elements are all starting from zero. By the definition in  \cref{eq:Kronecker}, the element in the coupling attention matrix $\widehat{AM}$ can be written as:
\begin{equation}
\label{eq:element_eq}
    am_{ij} = a_{[i//w,  j//w]} \times b_{[i\%w,  j\%w]}
\end{equation}

Here $am_{ij}$ denotes the element in the coupling attention matrix $\widehat{AM}$, which represents the alignment score of any two tokens from $Q$ and $K$. The subscript of $am_{ij}$ denotes the 1D indexes $i$ and $j$ of given tokens. The regular expansion from 2D feature map to 1D tokens can be written as follow: 

\begin{equation}
\label{eq:pos_to_index}
\left\{
\begin{aligned}
    &i = x^{(i)} + y^{(i)}* w\\
    &j = x^{(j)} + y^{(j)}* w
\end{aligned}
\right.
\end{equation}

\begin{figure*}
\centering  

\begin{subfigure}[b]{0.4\textwidth}
         \centering
         \includegraphics[width=\textwidth]{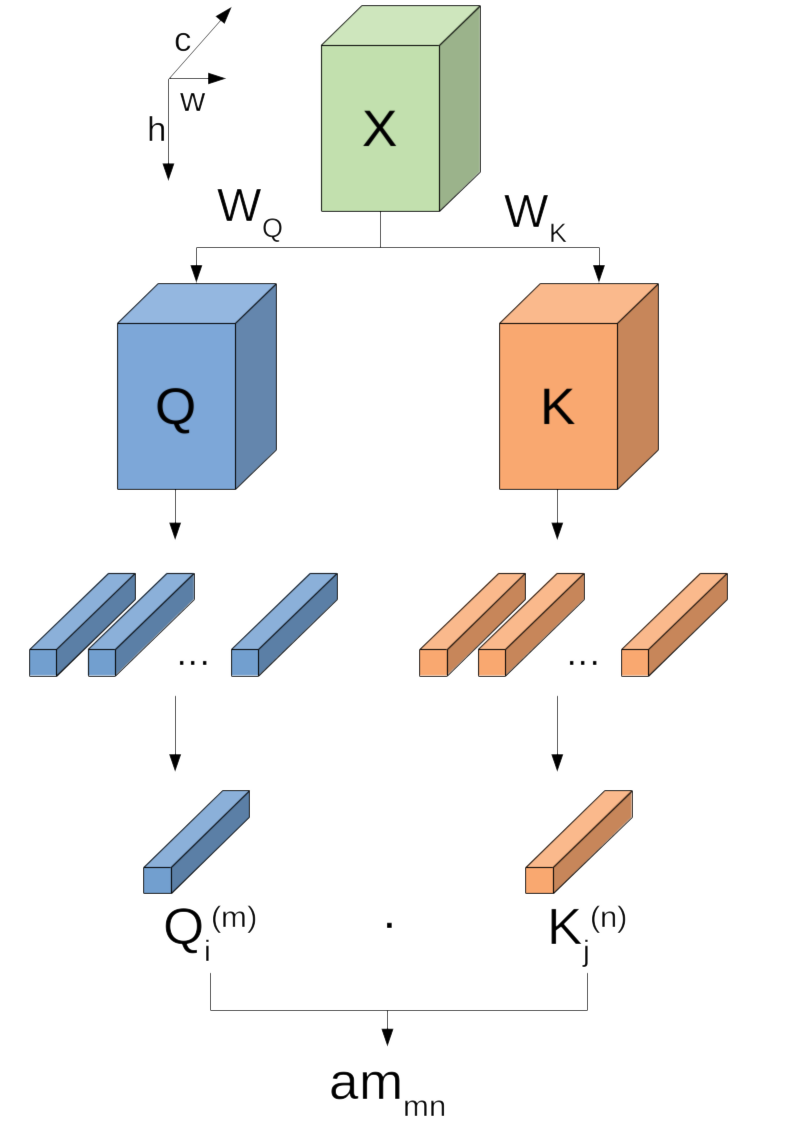}
         \caption{standard attention mechanism}
         \label{fig:vanilla_attention}
     \end{subfigure}
     \begin{subfigure}[b]{0.4\textwidth}
         \centering
         \includegraphics[width=\textwidth]{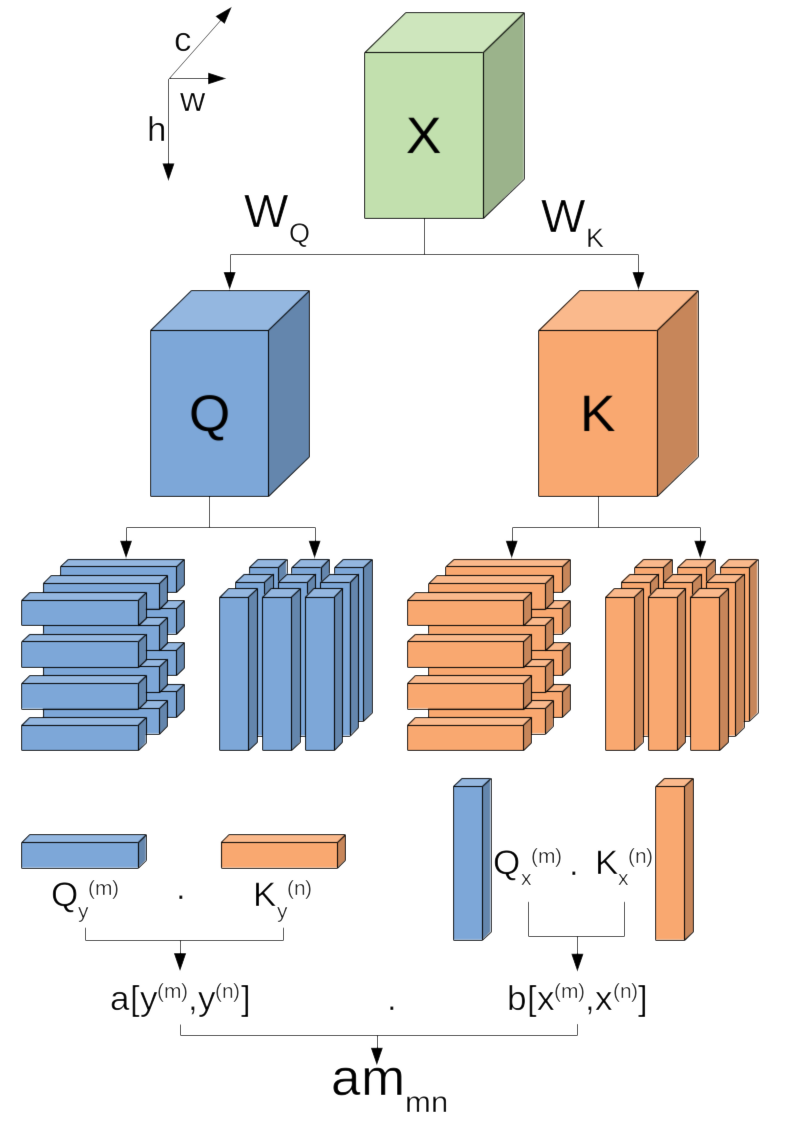}
         \caption{Equivalence attention mechanism by Kronecker Product}
         \label{fig:our_attention}
     \end{subfigure}
\caption{Left: standard attention mechanism, using the dot product between two vectors along the channel axis to obtain the attention map. Right: Our attention mechanism, using the product of two alignment scores of vectors along the height and width axis to obtain the attention map.}
\label{fig:attention_mechanism}
\end{figure*}

The $x^{(i)}$ and $y^{(i)}$ are the 2D spatial position of the i-th token. Then we substitute the $i$ and $j$ in \cref{eq:element_eq} with \cref{eq:pos_to_index}, the element of coupling attention matrix can be represented as:

\begin{equation}
\label{eq:element_pos}
    am_{ij} = a_{[y^{(i)},  y^{(j)}]} \times b_{[x^{(i)},  x^{(j)}]}
\end{equation}

Based on the hypothesis~\ref{hyp:appro} and \cref{eq:element_pos}, we construct the sub-matrix $A$ and $B$ from a novel perspective of attention mechanism. Give two points $\textbf{m}$ and $\textbf{n}$ on the 2D feature map with spatial position $(x^{(m)},y^{(m)})$ and $(x^{(n)},y^{(n)})$, as the \cref{fig:attention_mechanism} show, in the standard attention mechanism, the alignment score $am_{ij}$ of point $m$ and $n$ is calculated by the dot product between the tokens $Q_i^{(m)}$ and $K_j^{(n)}$ which are the vectors among the channel direction. To construct the sub-matrix $A \in \mathbb{R}^{h \times h}$ and $B \in \mathbb{R}^{w \times w}$, we change the way of expansing input tensor for dot product, as the \cref{fig:our_attention} shows, it is natural to consider $a_{[y^{(i)},  y^{(j)}]}$ as the alignment score between the row vectors along the width axis of point $m$ and $n$, while the $b_{[x^{(i)},  x^{(j)}]}$ can be seen as the alignment score between the column vectors. According the \cref{eq:element_pos}, then we can replace the original alignment score $am_{ij}$ with the product of $a_{[y^{(i)},  y^{(j)}]}$ and $b_{[x^{(i)},  x^{(j)}]}$. Thus, the intuition of \cref{eq:element_pos} is, under the hypothesis~\ref{hyp:appro}, that if two sub-matrix $A \in \mathbb{R}^{h \times h}$ and $B \in \mathbb{R}^{w \times w}$ can approximately couple the $AW \in \mathbb{R}^{hw \times hw}$ by tensor product, the original alignment score of vectors along the channel dimension can convert to the product of two alignment scores of vectors along the width and height dimension respectively. Therefore, the coupling attention mechanism can be regarded as capturing the similarity on the 2D feature map with the spatial information rather than the channel-wise information. Furthermore, coupling attention mechanism can be efficiently calculated by the $vec$ operator trick which is described in detail in the next subsection. 


\subsection{Efficient Calculation}
\label{sec:0303}
The calculating procedure of coupling attention mechanism can be simplified to significantly reduce the time and space complexity of algorithm, the output $O$ in the \cref{eq:vanilla_attention} can be obtained without explicitly calculating the $\widehat{AM}$ by vectorization trick.

\begin{lemma}
\label{lemma1}
The Kronecker product can be used to get a convenient representation for some matrix equations:

$$ (A \otimes B) \cdot row(X) = row(A\cdot X \cdot B^\top) $$

Here $row(\cdot)$ denotes the vectorization of the matrix $X$, which stack the rows of a matrix $X \in \mathbb{R}^{m \times n} $ one underneath the other to obtain a single vector $row(X) \in \mathbb{R}^{mn}$.

\end{lemma}
In the equation \ref{eq:vanilla_attention}, the output of original attention mechanism $O$ is the dot product between the softmax of attention matrix and the $V$. With the lemma\ref{lemma1}, the equation \ref{eq:vanilla_attention} can be written as below:

\begin{equation}
\label{eq:efficient_attention}
    SM(\widehat{AM}) \cdot row(V) \approx row(SM(A)\cdot V \cdot SM(B)^\top)
\end{equation}

In \cref{eq:efficient_attention} $V$ keep the shape $\mathbb{R}^{c \times h \times w}$ and is applied the matrix multiplication with $A$ and $B^\top$ with shape $\mathbb{R}^{c \times w \times w}$ and $\mathbb{R}^{c \times h \times h}$ in succession. Moreover, we empirically split out the softmax operation on sub-matrix $A$ and $B$, respectively. In the implementation of algorithm, we maintain the concept of multi-head structure to keep the same form with standard Transformer. And we find that the performance of model is sensitive with the number of heads, more detail will be discussed in the \cref{sec:04}.

\paragraph{Complexity Analysis} To better illustrate the efficiency  of our model, we provide some complexity analysis in this section. As described in the Section~\ref{sec:0303}, $L = h*w$. According to the \cref{eq:efficient_attention}, the computation complexity of the coupled attention block is $\mathcal{O}(2hw)$, which is less than the $\mathcal{O}((hw)^{2})$ in standard self-attention. In the aspect of the number of training parameters, our model has $4(hw)d$ training parameters, which is the same as standard self-attention. About the FLOPS(floating-point operation), depending on the calculation equation introduced in\cite{floating}, our model is less $4(hw)^{2}d+(hw)(4-(hw)-d-8\sqrt{(hw)})$ FLOPs than the standard Visual Transformer in each self-attention block. 

\subsection{Model Structure}
\label{sec:0304}
\begin{figure*}[ht!]
  \centering
    \begin{subfigure}{0.53\textwidth}
      \centering   
      \includegraphics[width=1\linewidth]{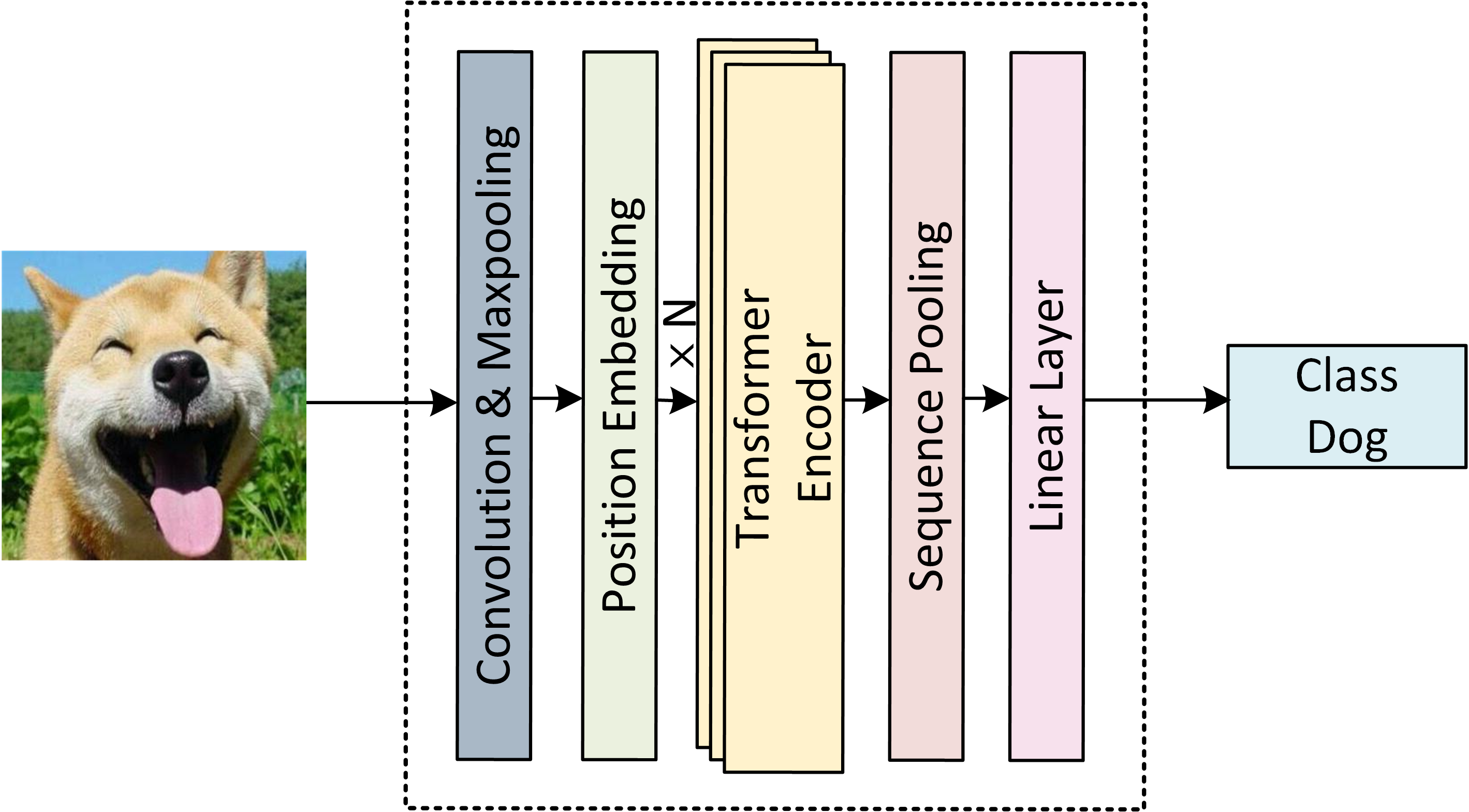}
        \caption{Architecture}
        \label{fig:sub1}
    \end{subfigure}   %
    \begin{subfigure}{0.43\textwidth}
      \centering   
      \includegraphics[width=\linewidth]{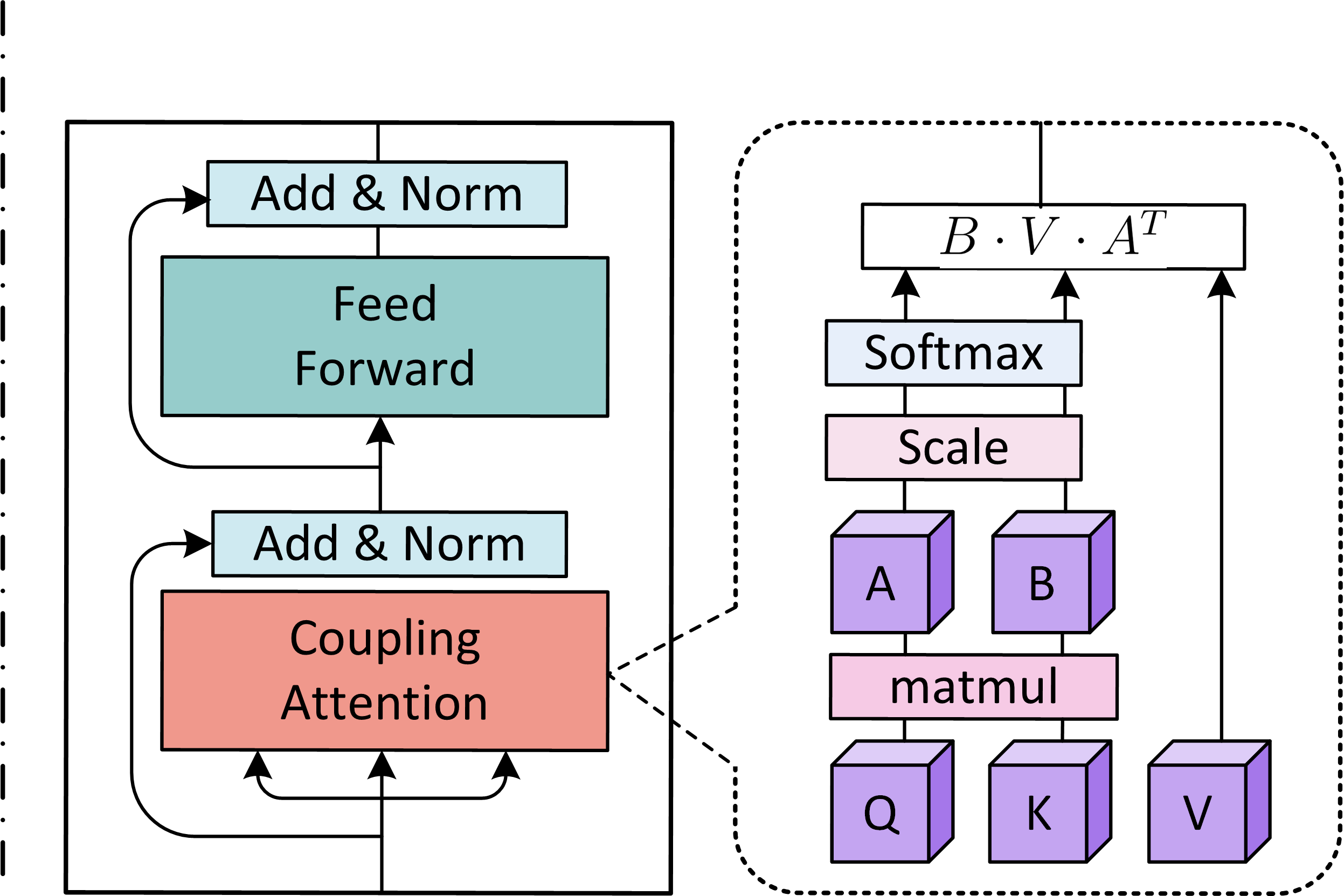}
        \caption{Couplformer block}
        \label{fig:sub2}
    \end{subfigure}
\caption{
\label{fig:architecture}
Illustrating the Couplformer architecture.
}
\end{figure*}

To adopt the Couplformer for image classification, we are inspired by \cite{levit,cct} and elaborate an architecture based on the ViT\cite{ViT16times16}. In this section, we would like to introduce the structures specifically. The total framework of our model is presented in the \cref{fig:architecture}.  

\paragraph{Patch Embedding} In standard Vision Transformer, the images are reshaped into a sequence of flattened 2D patches by adopting a trained linear projection\cite{ViT16times16}. Recently, the experiment demonstrates that the convolutional stem performs better than the patch stem\cite{cct,levit,DBLP:journals/corr/abs-2106-14881}. Therefore, we employ the convolutional layer before Transformer blocks. 

\paragraph{Position Embedding} In a standard Transformer, position embedding is an essential way regarding encoding spatial information\cite{ViT16times16}. Following the mainstream of the Transformer researcher, we also add the learned positional embedding layer to capture the relative distance between patches. However, due to the spatial information capture ability of Couplformer, we infer that the position embedding plays a less important role in Couplformer than standard Transformer, and the experiment results in \cref{sec:04} prove this inference.  

\paragraph{Encoder with Coupling Attention} The block of the encoder consists of two parts: the multi-head attention block and the feed-forward network. In our architecture, we replace the original multi-head attention block with coupling attention which can capture spatial information in visual tasks and release the memory consumption. Moreover, the other components, such as the Layer Normalization (LN) and Feedforward Network (FFN), are kept the same with the regular encoder. 

\paragraph{Classification Token} According to the standard approach of the Transformer framework, an extra learnable classification token is added to the sequence. However, in our model, we want to keep the shape of input feature maps in the Transformer encoder. Therefore, we apply the sequence pooling structure\cite{cct} to remove the extra classification token by which the desired sequence length is destroyed. In this structure, the model could distribute more weight to the patch, which contains more information relevant to the classifier. 


\begin{table*}[t]
\centering
\begin{tabular}{c|ccc|cc}
\toprule
Model &	CIFAR-10 & CIFAR-100 & MNIST & Params & MACs 	\\
\hline
\multicolumn{6}{c}{Convolutional Networks}\\
\hline
\textbf{ResNet34\cite{Resnet}}& 89.45\% & 64.67\% & 99.36\% & 21.80M & 0.09G\\
\textbf{ResNet50\cite{Resnet}}& 89.30\% & 61.25\% & 99.12\% & 25.56M & 0.09G\\
\textbf{MobileNet\cite{Mobilenet}}& 90.55\% & 67.12\% & 99.38\% & 8.72M & \textbf{0.03G}\\
\hline
\multicolumn{6}{c}{Vision Transformer}\\
\hline
\textbf{ViT-Base}\cite{ViT16times16} & 76.42\% & 46.61\% & 82.66\% & 85.63M & 0.43G \\
\textbf{ViT-Lite-7$/$4}\cite{cct} & 91.38\% & 69.74\% & 99.29\% & 3.72M & 0.24G   \\
\hline
\multicolumn{6}{c}{Efficient Transformer}\\
\hline
\textbf{Backbone-7/3$\times$2}\cite{cct} & \textbf{93.65\%} & 74.52\% & 99.26\% & 3.85M & 0.28G   \\
\textbf{Linformer\dag}\cite{linformer} & 92.45\% & 70.87\% & 99.13\% & 3.96M & 0.28G    \\
\textbf{Performer}\cite{performer} & 91.58\% & 73.11\% & \textbf{99.29\%} & 3.85M & 0.28G\\
\textbf{Reformer}\cite{reformer} & 90.58\% & 73.02\% & 99.11\% & 3.39M & 0.25G\\
\textbf{Couplformer-4} & 90.81\% & 69.19\% & 99.12\% & \textbf{0.48M} & 0.10G  \\
\textbf{Couplformer-7} & 93.44\% & \textbf{74.53\%} & 99.14\% & 3.85M & 0.28G  \\
\textbf{Couplformer-14} & 92.15\% & 67.22\% & - & 20.94M & 1.38G  \\
\textbf{Couplformer-14*} & - & 74.61\% & - & 20.94M & 1.38G  \\
\bottomrule
    \end{tabular}
    \caption{Top-1 validation accuracy comparisons. MACs means the Multiply-Accumulate operations. ViT-Lite-7/4 denote the model consists of 7 layer Transformer encoders with 4 $\times$ 4 patches. Backbone-7/3$\times$2 denotes the model consists of 7 layer Transformer encoders and 2 Convolution layers with 3$\times$3 kernal size. \dag indicates the Linformer model reduce the sequence length to half. * indicates training with pre-trained model}
    \label{tab:cifar-mnist}
\end{table*}
\begin{table*}
\centering
\begin{tabular}{c|c|cc|c}
\toprule
Model &	 ImageNet-1k & Memory & Embed dim/heads & Training Epochs 	\\
\hline
\textbf{Backbone-14} & 67.72\%(baseline) & 16421M(baseline) & 256/4 & 50\\
\hline
 & 66.21\%(-1.51\%) & 12701M(77.34\%) & 256/256 & 50\\
\textbf{Couplformer-14} & 66.22\%(-1.50\%) & 11949M(72.76\%) & 256/128 & 50\\
 & 62.22\%(-5.50\%) & \textbf{11503M}(70.05\%) & 256/64 & 50\\
\hline
 & 67.96\%(+0.24\%) & 17141M(104.38\%) & 384/384 & 50\\
\textbf{Couplformer-14} & 68.34\%(+0.62\%) & 16259M(99.01\%) & 384/192 & 50\\
 & \textbf{68.64\%}(+0.92\%) & 15379M(93.65\%) & 384/96 & 50\\

\bottomrule
    \end{tabular}
    \caption{ImageNet: Top-1 validation accuracy comparisons under first 50 training epochs. Embed dim means the Embedding dimension. }
    \label{tab:imagenet}
\end{table*}

\section{Experiments}
\label{sec:04}
In this section, we investigate the model capabilities of Coulpformer, ConvNets, and other Vision Transformers on different scales of image classification tasks. Then, considering the utilization of spatial information in the coupling attention mechanism, we discuss the necessity of importing position embedding. Lastly, we evaluate the performance of Coulpformer when pre-trained on a large-scale dataset.
\subsection{Setup}
\paragraph{Dataset}
In order to explore the scalability of Couplformer, four benchmark datasets  MNIST\cite{MNIST}, CIFAR-10, CIFAR-100, and ImageNet-1k\cite{imagenet} are used for different purpose. Among those datasets, MNIST, CIFAR-10, and CIFAR-100 represent the small-sized datasets, while ImageNet-1k represents the medium-sized. 

\paragraph{Implementation Details}
In our experiments, all kinds of efficient Transformer approaches are evaluated on the same backbone framework\cite{cct} and report the best results we could obtain. We select top-1\% accuracy as evaluation metrics. We train our model with a single GeForce RTX 2080Ti(11GB) for small-sized datasets and use a server with 8 GeForce RTX 3090(24GB) for training on a medium-sized dataset.  

\begin{table*}
\centering
\begin{tabular}{c|c|c}
\toprule
Model &	 number of heads & CIFAR-100 	\\
\hline
 & 128 & 66.12\%(baseline) \\
 & 64 & 67.85\%(+1.73\%) \\
\textbf{Couplformer-7}& 32 & \textbf{73.93\%}(+7.81\%)\\
& 16 & 72.75\%(+6.63\%) \\
 & 8 & 68.20\%(+2.08\%)\\
\bottomrule
    \end{tabular}
    \caption{The comparison of Top-1 validation accuracy under different numbers of heads in multi-head self-attention.}
    \label{tab:number_head}
\end{table*}

\begin{table*}
\centering
\begin{tabular}{c|c|cc}
\toprule
Model &	 Pos Emb & CIFAR-10 & CIFAR-100 	\\
\hline
\textbf{ViT-12/16}\cite{ViT16times16} & None & 66.71\%(baseline) & 38.22\%(baseline)\\
& Learnable & 69.82\%(+3.11\%) & 40.57\%(+2.35\%)\\
\hline
\textbf{Backbone}\cite{cct} & None & 92.79\%(baseline) & 70.31\%(baseline)\\
& Learnable & 93.65\%(+0.86\%) & 74.52\%(+4.21\%)\\

\hline
\textbf{Couplformer-7} & None & 94.06\%(baseline) & 72.74\%(baseline)\\
& Learnable & 93.14\%(-0.92\%) & 73.98\%(+1.24\%)\\

\bottomrule
    \end{tabular}
    \caption{The comparison of Top-1 validation accuracy under different modes of position embedding.}
    \label{tab:position_embedding}
\end{table*}
\subsection{Image Classification}
For small-sized datasets, models are trained for 200 epochs. The learning rate is 3e-4, weight decay is 3e-2, and batch size is 128. For ImageNet-1k, we employ the learning rate of 5e-4, weight decay of 5e-2, batch size of 1024, and train for 50 epochs. All the training employ optimizer AdamW employing cosine decay learning rate scheduler with 10 epochs' linear warmup.

\paragraph{Evaluation in Small Datasets} As described in \cref{tab:cifar-mnist}, We conduct the image classification task in MNIST, CIFAR-10, and CIFAR100 datasets. We compare Couplformer with the standard Visual Transformer\cite{ViT16times16} and several popular efficient Transformers\cite{linformer,reformer,performer} on the same backbone\cite{cct} with 2 convolutional layers of $3\times3$ filter size. We also list two ConvNets baseline: ResNet\cite{Resnet} and MobileNet\cite{Mobilenet} for comparison. 

From the results in \cref{tab:cifar-mnist}, our model notably surpass the other efficient Transformers with similar complexity on most tasks. Compared with the backbone, Couplformer typically achieves the scarcely distinguishable accuracy with slight declining the memory usage, which is not obviously in processing the small resolution images. Additionally, Couplformer still outperforms the ConvNets without a significant increase in MACs. These experiments prove that our model could keep remarkable outcomes on small-scale datasets. 
\paragraph{Evaluation in ImageNet Datasets} \cref{tab:imagenet} summarizes the result of the evaluation on ImageNet-1k dataset. The input images are all resized to $224^2$. Compared with the backbone model, Couplformer reduces 27.24\% to 22.66\% of the memory consumption and reaches the same accuracy. Under the same level of memory consumption, our model brings up 0.92\% gains in Top-1 accuracy compared with backbone.

\paragraph{Scaling Study of Heads}
The model complexity is a positive correlation with the number of heads in the multi-head structure. From the experiments, we find the result is somewhat counterintuitive that the larger number of heads will not always improve the performance. Specifically, as presented in \cref{tab:number_head}, in the CIFAR100 dataset, the accuracy achieves the best performance only when the head number is 32, and the accuracy would fall no matter the number of heads increased or decreased. Besides, in the \cref{tab:imagenet}, Couplformer-14 trained on ImageNet-1k gets the highest results while setting the number of heads to 128/96 in 256/384 embedding dimension. These results illustrate the sensitivity of the number of heads.

\begin{figure}[t]
	\centering
    \includegraphics[width=0.48\textwidth]{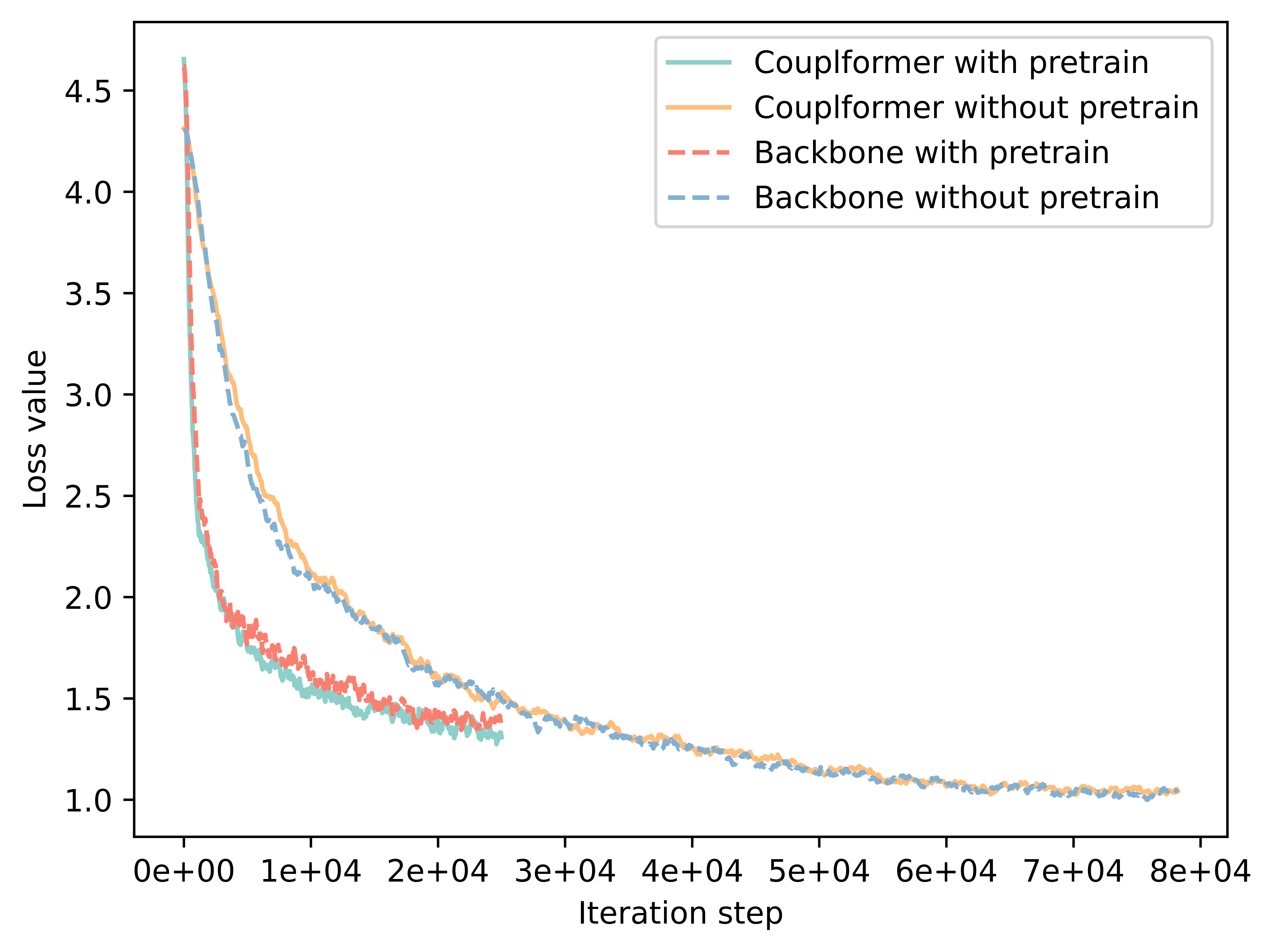}
	\caption{The loss value comparison about the pre-train transfer to CIFAR-100}
	\label{fig:pretrain}
\end{figure}
\subsection{Ablation Study}
\paragraph{Evaluation of Position Embedding}
We conduct the ablation study about the purpose of position embedding. In the Visual Transformer structure, adding learned positional embeddings to inputs has become mainstream. The position embedding has been proved that it could bring considerable accuracy improvement. However, in our model, the utilization of spatial information to generate the alignment scores in coupling the attention mechanism makes the positional embedding less important. \cref{tab:position_embedding} summarizes of the result from the investigation on ViT-12/16\cite{ViT16times16}, Backbone\cite{cct}, and Couplformer. As we see, the regular Transformer in ViT-12/16 and backbone are both benefited from position embedding to obtain a gap of improvement than the baseline without embedding position. In contrast, Couplformer appears unaffected with or without position embedding. This result proves that our model typically doesn't rely on position embedding. Besides, it also demonstrates that spatial information is implicitly employed, as we declare in \cref{sec:3}.

\subsection{Evaluation of Pre-training Performance}
In \cref{fig:pretrain}, we present the performance of our model pre-trained on ImageNet-1k. With the pre-trained model, both Couplformer and backbone attain $+0.08\%$ and $+0.31\%$ on CIFAR-100 by merely training for 50 epochs. It shows that our model shared the same ability to obtain excellent results when pre-trained at a larger-scale and transferred to a  smaller dataset. Moreover, in \cref{tab:cifar-mnist}, the Couplformer-L obtains a $+7.39\%$ accuracy gain with pre-training. 

\section{Conclusion}
In this paper, we presented a novel memory economy attention mechanism, named Couplformer, which employs spatial information to couple the attention map to replace the traditional self-attention module. Through this novel approach, the shortage of dramatic memory consumption of Visual Transformer is efficiently mitigated. We demonstrate that our model achieves sufficient accuracy requirements with the minimal occupation of GPU. Moreover, we apply spatial information to generate the alignment scores rather than channel-wise as the standard Visual Transformer model, and the experiments confirm the effectiveness of this coupling attention mechanism. Our model makes the Transformer model more flexible for different scales of defined settings. In the days of data explosion, our model helps more researchers who are suffering from limited hardware resources to train on the various sizes of datasets lightheartedly. We believe that our model will bring researchers a fresh perspective of deep learning architectures. 
%


{\small

\bibliographystyle{ieee_fullname}
\normalem
\bibliography{egbib}

\begin{thebibliography}{10}\itemsep=-1pt

\bibitem{bello2019attention_augment_iccv}
Irwan Bello, Barret Zoph, Ashish Vaswani, Jonathon Shlens, and Quoc~V Le.
\newblock Attention augmented convolutional networks.
\newblock In {\em Proceedings of the IEEE International Conference on Computer
  Vision (ICCV)}, pages 3286--3295, 2019.

\bibitem{longformer}
Iz Beltagy, Matthew~E Peters, and Arman Cohan.
\newblock Longformer: The long-document transformer.
\newblock {\em arXiv preprint arXiv:2004.05150}, 2020.

\bibitem{lowrank}
Srinadh Bhojanapalli, Chulhee Yun, Ankit~Singh Rawat, Sashank Reddi, and Sanjiv
  Kumar.
\newblock Low-rank bottleneck in multi-head attention models.
\newblock In {\em International Conference on Machine Learning(ICML)}, pages
  864--873, 2020.

\bibitem{endtoendOD}
Nicolas Carion, Francisco Massa, Gabriel Synnaeve, Nicolas Usunier, Alexander
  Kirillov, and Sergey Zagoruyko.
\newblock End-to-end object detection with transformers.
\newblock In {\em 16th European Conference of Computer Vision - {ECCV},
  Glasgow, UK, August 23-28, Proceedings, Part {I}}, pages 213--229, 2020.

\bibitem{chen2017sca_cvpr}
Long Chen, Hanwang Zhang, Jun Xiao, Liqiang Nie, Jian Shao, Wei Liu, and
  Tat-Seng Chua.
\newblock Sca-cnn: Spatial and channel-wise attention in convolutional networks
  for image captioning.
\newblock In {\em Proceedings of the IEEE conference on computer vision and
  pattern recognition(CVPR)}, pages 5659--5667, 2017.

\bibitem{igpt}
Mark Chen, Alec Radford, Jeff Wu, Heewoo Jun, Prafulla Dhariwal, David Luan,
  and Ilya Sutskever.
\newblock Generative pretraining from pixels.
\newblock In {\em Proceedings of the 37th International Conference on Machine
  Learning(ICML)}, 2020.

\bibitem{sparformer}
Rewon Child, Scott Gray, Alec Radford, and Ilya Sutskever.
\newblock Generating long sequences with sparse transformers.
\newblock {\em ArXiv}, abs/1904.10509, 2019.

\bibitem{performer}
Krzysztof~Marcin Choromanski, Valerii Likhosherstov, David Dohan, Xingyou Song,
  Andreea Gane, Tam{\'{a}}s Sarl{\'{o}}s, Peter Hawkins, Jared~Quincy Davis,
  Afroz Mohiuddin, Lukasz Kaiser, David~Benjamin Belanger, Lucy~J. Colwell, and
  Adrian Weller.
\newblock Rethinking attention with performers.
\newblock In {\em 9th International Conference on Learning Representations,
  {(ICLR)}, Virtual Event, Austria, May 3-7}, 2021.

\bibitem{CPVT}
Xiangxiang Chu, Bo Zhang, Zhi Tian, Xiaolin Wei, and Huaxia Xia.
\newblock Do we really need explicit position encodings for vision
  transformers?
\newblock {\em arXiv e-prints}, pages arXiv--2102, 2021.

\bibitem{Cordonnier2020On_relation_iclr}
Jean-Baptiste Cordonnier, Andreas Loukas, and Martin Jaggi.
\newblock On the relationship between self-attention and convolutional layers.
\newblock In {\em International Conference on Learning Representations (ICLR)},
  2020.

\bibitem{transX}
Zihang Dai, Zhilin Yang, Yiming Yang, Jaime~G. Carbonell, Quoc~Viet Le, and
  Ruslan Salakhutdinov.
\newblock Transformer-xl: Attentive language models beyond a fixed-length
  context.
\newblock In {\em Proceedings of the 57th Conference of the Association for
  Computational Linguistics, {ACL}, Florence, Italy}, 2019.

\bibitem{imagenet}
Jia Deng, Wei Dong, Richard Socher, Li-Jia Li, Kai Li, and Li Fei-Fei.
\newblock Imagenet: A large-scale hierarchical image database.
\newblock In {\em In Proceedings of the IEEE conference on computer vision and
  pattern recognition}, pages 248--255, 2009.

\bibitem{ViT16times16}
Alexey Dosovitskiy, Lucas Beyer, Alexander Kolesnikov, Dirk Weissenborn,
  Xiaohua Zhai, Thomas Unterthiner, Mostafa Dehghani, Matthias Minderer, Georg
  Heigold, Sylvain Gelly, Jakob Uszkoreit, and Neil Houlsby.
\newblock An image is worth 16x16 words: Transformers for image recognition at
  scale.
\newblock In {\em 9th International Conference on Learning Representations,
  {(ICLR)}, Virtual Event, Austria, May 3-7}, 2021.

\bibitem{permuteequivaraint}
Vijay~Prakash Dwivedi and Xavier Bresson.
\newblock A generalization of transformer networks to graphs.
\newblock {\em arXiv preprint arXiv:2012.09699}, 2020.

\bibitem{FAHIM2020106437}
Shahriar~Rahman Fahim, Yeahia Sarker, Subrata~K Sarker, Md~Rafiqul~Islam
  Sheikh, and Sajal~K Das.
\newblock Self attention convolutional neural network with time series imaging
  based feature extraction for transmission line fault detection and
  classification.
\newblock {\em Electric Power Systems Research}, 187:106437, 2020.

\bibitem{dualatt}
J. Fu, J. Liu, Haijie Tian, Zhiwei Fang, and Hanqing Lu.
\newblock Dual attention network for scene segmentation.
\newblock {\em 2019 IEEE/CVF Conference on Computer Vision and Pattern
  Recognition (CVPR)}, pages 3141--3149, 2019.

\bibitem{cnndeeplearn}
Ian Goodfellow, Yoshua Bengio, and Aaron Courville.
\newblock {\em Deep learning}.
\newblock MIT press, 2016.

\bibitem{levit}
Ben Graham, Alaaeldin El-Nouby, Hugo Touvron, Pierre Stock, Armand Joulin,
  Herv{\'e} J{\'e}gou, and Matthijs Douze.
\newblock Levit: a vision transformer in convnet's clothing for faster
  inference.
\newblock {\em arXiv preprint arXiv:2104.01136}, 2021.

\bibitem{cct}
Ali Hassani, Steven Walton, Nikhil Shah, Abulikemu Abuduweili, Jiachen Li, and
  Humphrey Shi.
\newblock Escaping the big data paradigm with compact transformers.
\newblock {\em arXiv preprint arXiv:2104.05704}, 2021.

\bibitem{Resnet}
Kaiming He, Xiangyu Zhang, Shaoqing Ren, and Jian Sun.
\newblock Deep residual learning for image recognition.
\newblock In {\em Proceedings of the IEEE conference on computer vision and
  pattern recognition(CVPR)}, pages 770--778, 2016.

\bibitem{axialtransformer}
Jonathan Ho, Nal Kalchbrenner, Dirk Weissenborn, and Tim Salimans.
\newblock Axial attention in multidimensional transformers.
\newblock {\em ArXiv}, abs/1912.12180, 2019.

\bibitem{relationOD}
Han Hu, Jiayuan Gu, Zheng Zhang, Jifeng Dai, and Yichen Wei.
\newblock Relation networks for object detection.
\newblock In {\em In processing of the {IEEE} Conference on Computer Vision and
  Pattern Recognition, {CVPR}, Salt Lake City, UT, USA, June 18-22, 2018},
  2018.

\bibitem{gatherexcite}
Jie Hu, Li Shen, Samuel Albanie, Gang Sun, and Andrea Vedaldi.
\newblock Gather-excite: Exploiting feature context in convolutional neural
  networks.
\newblock In {\em Advances in Neural Information Processing Systems 31: Annual
  Conference on Neural Information Processing Systems 2018(NeurIPS), December,
  3-8, Montr{\'{e}}al, Canada}, 2018.

\bibitem{SEnet}
Jie Hu, Li Shen, Samuel Albanie, Gang Sun, and Enhua Wu.
\newblock Squeeze-and-excitation networks.
\newblock {\em IEEE Transactions on Pattern Analysis and Machine Intelligence},
  42:2011--2023, 2020.

\bibitem{huang2019attention_AOA_iccv}
Lun Huang, Wenmin Wang, Jie Chen, and Xiao-Yong Wei.
\newblock Attention on attention for image captioning.
\newblock In {\em Proceedings of the IEEE International Conference on Computer
  Vision (ICCV)}, pages 4634--4643, 2019.

\bibitem{floating}
Raphael Hunger.
\newblock {\em Floating point operations in matrix-vector calculus}.
\newblock Munich University of Technology, Inst. for Circuit Theory and Signal,
  2005.

\bibitem{linearTransformer}
Angelos Katharopoulos, Apoorv Vyas, Nikolaos Pappas, and Fran{\c{c}}ois
  Fleuret.
\newblock Transformers are rnns: Fast autoregressive transformers with linear
  attention.
\newblock In {\em International Conference on Machine Learning(ICML)}, pages
  5156--5165, 2020.

\bibitem{vitsurvey}
Salman Khan, Muzammal Naseer, Munawar Hayat, Syed~Waqas Zamir, Fahad~Shahbaz
  Khan, and Mubarak Shah.
\newblock Transformers in vision: A survey.
\newblock {\em arXiv preprint arXiv:2101.01169}, 2021.

\bibitem{reformer}
Nikita Kitaev, Lukasz Kaiser, and Anselm Levskaya.
\newblock Reformer: The efficient transformer.
\newblock In {\em 8th International Conference on Learning Representations,
  {ICLR}, Addis Ababa, Ethiopia, April 26-30, 2020}, 2020.

\bibitem{MNIST}
Yann LeCun, L{\'e}on Bottou, Yoshua Bengio, and Patrick Haffner.
\newblock Gradient-based learning applied to document recognition.
\newblock {\em Proceedings of the IEEE}, 86(11):2278--2324, 1998.

\bibitem{swintrans}
Ze Liu, Yutong Lin, Yue Cao, Han Hu, Yixuan Wei, Zheng Zhang, Stephen Lin, and
  Baining Guo.
\newblock Swin transformer: Hierarchical vision transformer using shifted
  windows.
\newblock In {\em Proceedings of the IEEE/CVF International Conference on
  Computer Vision (ICCV)}, pages 10012--10022, October 2021.

\bibitem{locality}
Niki Parmar, Ashish Vaswani, Jakob Uszkoreit, Lukasz Kaiser, Noam Shazeer,
  Alexander Ku, and Dustin Tran.
\newblock Image transformer.
\newblock In {\em International Conference on Machine Learning(ICML)}, pages
  4055--4064, 2018.

\bibitem{blockwise}
Jiezhong Qiu, Hao Ma, Omer Levy, Wen{-}tau Yih, Sinong Wang, and Jie Tang.
\newblock Blockwise self-attention for long document understanding.
\newblock In {\em Findings of the Association for Computational Linguistics:
  {EMNLP} Online Event, 16-20 November}, 2020.

\bibitem{CompreTrans}
Jack~W. Rae, Anna Potapenko, Siddhant~M. Jayakumar, Chloe Hillier, and
  Timothy~P. Lillicrap.
\newblock Compressive transformers for long-range sequence modelling.
\newblock In {\em 8th International Conference on Learning Representations,
  {ICLR} , Addis Ababa, Ethiopia, April 26-30,}, 2020.

\bibitem{standalone}
Prajit Ramachandran, Niki Parmar, Ashish Vaswani, Irwan Bello, Anselm Levskaya,
  and Jonathon Shlens.
\newblock Stand-alone self-attention in vision models.
\newblock In {\em NeurIPS}, 2019.

\bibitem{Mobilenet}
Mark Sandler, Andrew Howard, Menglong Zhu, Andrey Zhmoginov, and Liang-Chieh
  Chen.
\newblock Mobilenetv2: Inverted residuals and linear bottlenecks.
\newblock In {\em Proceedings of the IEEE conference on computer vision and
  pattern recognition}, pages 4510--4520, 2018.

\bibitem{synthesizer}
Yi Tay, Dara Bahri, Donald Metzler, Da-Cheng Juan, Zhe Zhao, and Che Zheng.
\newblock Synthesizer: Rethinking self-attention for transformer models.
\newblock In {\em International Conference on Machine Learning}, pages
  10183--10192. PMLR, 2021.

\bibitem{efficientsurvey}
Yi Tay, Mostafa Dehghani, Dara Bahri, and Donald Metzler.
\newblock Efficient transformers: A survey.
\newblock {\em arXiv preprint arXiv:2009.06732}, 2020.

\bibitem{mlpmixer}
Ilya Tolstikhin, Neil Houlsby, Alexander Kolesnikov, Lucas Beyer, Xiaohua Zhai,
  Thomas Unterthiner, Jessica Yung, Andreas Steiner, Daniel Keysers, Jakob
  Uszkoreit, et~al.
\newblock Mlp-mixer: An all-mlp architecture for vision.
\newblock {\em arXiv preprint arXiv:2105.01601}, 2021.

\bibitem{DeiT}
Hugo Touvron, Matthieu Cord, Matthijs Douze, Francisco Massa, Alexandre
  Sablayrolles, and Herv{\'e} J{\'e}gou.
\newblock Training data-efficient image transformers \& distillation through
  attention.
\newblock In {\em International Conference on Machine Learning}, pages
  10347--10357. PMLR, 2021.

\bibitem{attentionisallyouneed}
Ashish Vaswani, Noam Shazeer, Niki Parmar, Jakob Uszkoreit, Llion Jones,
  Aidan~N Gomez, {\L}ukasz Kaiser, and Illia Polosukhin.
\newblock Attention is all you need.
\newblock In {\em Advances in neural information processing systems}, pages
  5998--6008, 2017.

\bibitem{residualatt}
Fei Wang, Mengqing Jiang, Chen Qian, Shuo Yang, Cheng Li, Honggang Zhang,
  Xiaogang Wang, and Xiaoou Tang.
\newblock Residual attention network for image classification.
\newblock In {\em Proceedings of the IEEE Conference on Computer Vision and
  Pattern Recognition (CVPR)}, July 2017.

\bibitem{axialsegmentation}
Huiyu Wang, Yukun Zhu, Bradley Green, Hartwig Adam, Alan~Loddon Yuille, and
  Liang-Chieh Chen.
\newblock Axial-deeplab: Stand-alone axial-attention for panoptic segmentation.
\newblock In {\em ECCV}, 2020.

\bibitem{linformer}
Sinong Wang, Belinda~Z. Li, Madian Khabsa, Han Fang, and Hao Ma.
\newblock Linformer: Self-attention with linear complexity.
\newblock {\em ArXiv}, abs/2006.04768, 2020.

\bibitem{cbam}
Sanghyun Woo, Jongchan Park, Joon-Young Lee, and In~So Kweon.
\newblock Cbam: Convolutional block attention module.
\newblock In {\em Proceedings of the European Conference on Computer Vision
  (ECCV)}, September 2018.

\bibitem{CVT}
Haiping Wu, Bin Xiao, Noel Codella, Mengchen Liu, Xiyang Dai, Lu Yuan, and Lei
  Zhang.
\newblock Cvt: Introducing convolutions to vision transformers.
\newblock {\em arXiv preprint arXiv:2103.15808}, 2021.

\bibitem{DBLP:journals/corr/abs-2106-14881}
Tete Xiao, Mannat Singh, Eric Mintun, Trevor Darrell, Piotr Doll{\'{a}}r, and
  Ross~B. Girshick.
\newblock Early convolutions help transformers see better.
\newblock {\em CoRR}, 2021.

\bibitem{DBLP:conf/aaai/XiongZCTFLS21}
Yunyang Xiong, Zhanpeng Zeng, Rudrasis Chakraborty, Mingxing Tan, Glenn Fung,
  Yin Li, and Vikas Singh.
\newblock Nystr{\"{o}}mformer: {A} nystr{\"{o}}m-based algorithm for
  approximating self-attention.
\newblock In {\em Thirty-Fifth {AAAI} Conference on Artificial Intelligence},
  pages 14138--14148, 2021.

\bibitem{memformer}
Qing yang Wu, Zhenzhong Lan, Jing Gu, and Zhou Yu.
\newblock Memformer: The memory-augmented transformer.
\newblock {\em ArXiv}, 2020.

\bibitem{CeiT}
Kun Yuan, Shaopeng Guo, Ziwei Liu, Aojun Zhou, Fengwei Yu, and Wei Wu.
\newblock Incorporating convolution designs into visual transformers.
\newblock {\em arXiv preprint arXiv:2103.11816}, 2021.

\bibitem{bigbird}
Manzil Zaheer, Guru Guruganesh, Kumar~Avinava Dubey, Joshua Ainslie, Chris
  Alberti, Santiago Onta{\~{n}}{\'{o}}n, Philip Pham, Anirudh Ravula, Qifan
  Wang, Li Yang, and Amr Ahmed.
\newblock Big bird: Transformers for longer sequences.
\newblock In {\em Advances in Neural Information Processing Systems 33: Annual
  Conference on Neural Information Processing Systems, NeurIPS, December ,
  virtual}, 2020.

\bibitem{zhao2020exploring_CVPR}
Hengshuang Zhao, Jiaya Jia, and Vladlen Koltun.
\newblock Exploring self-attention for image recognition.
\newblock In {\em Proceedings of the IEEE/CVF Conference on Computer Vision and
  Pattern Recognition (CVPR)}, pages 10076--10085, 2020.

\bibitem{detectdeformable}
Xizhou Zhu, Weijie Su, Lewei Lu, Bin Li, Xiaogang Wang, and Jifeng Dai.
\newblock Deformable {DETR:} deformable transformers for end-to-end object
  detection.
\newblock In {\em 9th International Conference on Learning Representations,
  {ICLR} , Virtual Event, Austria, May 3-7}, 2021.

\end{thebibliography}
}


\end{document}